\documentclass[conference]{IEEEtran}
\usepackage{cite}
\usepackage{amsmath,amssymb,amsfonts}
\usepackage{algorithmic}
\usepackage{graphicx}
\usepackage{textcomp}
\usepackage{xcolor}
\usepackage{hyperref}
\def\BibTeX{{\rm B\kern-.05em{\sc i\kern-.025em b}\kern-.08em
    T\kern-.1667em\lower.7ex\hbox{E}\kern-.125emX}}
\begin{document}

\title{Image Segmentation and Classification of E-waste for Training Robots for Waste Segregation\\}

\author{\IEEEauthorblockN{Prakriti Tripathi}
\IEEEauthorblockA{
\textit{Indian Institute of Technology Dharwad}\\
Dharwad, India \\
220120014@iitdh.ac.in or prakrititripathi3@gmail.com}
}

\maketitle

\begin{abstract}
Industry partners provided a problem statement that involves classifying electronic waste using machine learning models, which will be utilized by pick-and-place robots for waste segregation. This was achieved by taking common electronic waste items, such as a mouse and a charger, unsoldering them, and taking pictures to create a custom dataset. The state-of-the-art YOLOv11 model was trained and run to achieve 70 mAP in real-time. The Mask R-CNN model was also trained and achieved 41 mAP. The model can be integrated with pick-and-place robots to perform segregation of e-waste. The code and link to the dataset can be found in this GitHub repository:
\href{https://tinyurl.com/5dnp4av5}{https://github.com/prakriti16/Image-segmentation-and-classification-of-e-waste}
\end{abstract}


\section{Introduction}
Electronic waste (e-waste) is one of the fastest-growing solid waste streams globally \cite{b2}. 
In the recycling plants in factories after the e-waste has been crushed several internal components such as resistors, capacitors and LEDs come out as is.
These components can be identified using our trained machine learning model, and then pick-and-place robots along the belt can segregate these items into various bins by size or type.

\section{Related Work}

A popular real time instance segmentation model on MS COCO dataset was RTMDet \cite{b4} which was based on YOLOv5 model and achieved 44.0 box AP and 38.7 mask AP with 10.18 M parameters. Recently, Ultralytics \cite{b1} released the YOLOv11 model, which achieves 46.6 box AP and 37.8 mask AP with 10.1 M parameters.

Specifically in the context of Trash detection, Mindy Yang and Gary Thung \cite{b3} had tried SVM and CNN models to classify trash into five categories and achieved 75\% accuracy on their custom dataset.

Pedro F. Proença  and Pedro Simões \cite{b5} used Mask R-CNN on the dataset they developed. They achieved a low mAP of 26.1 due to small objects with several instances being missed in the classification after the images were downsized. 

Niful Islam et al. \cite{b6} introduced EWasteNet, which is a data-efficient image transformer requiring just a hundred images to train. However, since it consists of a two-step process, it requires additional training time, making it unsuitable for real-time applications.

Most electronic waste datasets contain pictures of electronic equipments in a pristine and undamaged state. However, in recycling plants, after the crushing process, these items will be broken into tiny components. 

\section{Methodology}
\subsection{Dataset}
To create the dataset, some electronic devices were taken, like a mouse and a charger, their outer casing was removed, and the individual components inside were unsoldered. This simulates what happens after the electronic items have been crushed. 100 images of the individual components in all possible orientations were captured. A 22-second video was recorded of all the components placed near each other in a straight line to simulate the view from the belt drive in the recycling facility. This video was then sampled at 60 FPS to add images of individual frames to the dataset. An additional 100 images of these components were added from the internet to increase diversity in the dataset. 
 
Pre-processing included applying Auto-Orient, which ensures there is no misalignment. The images were divided into 2x2 tiles to make sure small components and details were captured. Data augmentation involved horizontal and vertical flipping, as well as rotation by 90 degrees, and adding 0.1\% noise perturbations to pixels to simulate camera quality.

The final dataset contained 6180 total images with 88\% training, 8\% validation, and 4\% test data. Each component in these images was assigned a category and annotated manually using an online tool \cite{b8}.

\subsection{Models}\label{AA}
\subsubsection{YOLOv11}
These images were passed through the YOLO11s-seg model, which contains 355 layers, 10,089,254 parameters, 10,089,238 gradients, and 35.6 GFLOPs. 
The architecture involves an initial sequential down-sampling of images by reducing the spatial dimensions (via convolution with stride 2), while later rows restore spatial resolution (via up-sampling and concatenation).
There are also some layers in between, like C3k2, SPPF, and C2PSA for efficient feature extraction and attention.

\subsubsection{Mask R-CNN}
Mask R-CNN consists of a ResNet Backbone to extract features, with deep layers capturing semantic information and shallow layers capturing fine details. FPN enhances the backbone by creating a feature pyramid, enabling the detection of objects at different scales. These are then passed through the Region Proposal Network and finally through the mask prediction heads to generate the final output.

\section{Results and Discussion}
\begin{figure}[htbp]
\centerline{\includegraphics[width=0.9\linewidth]{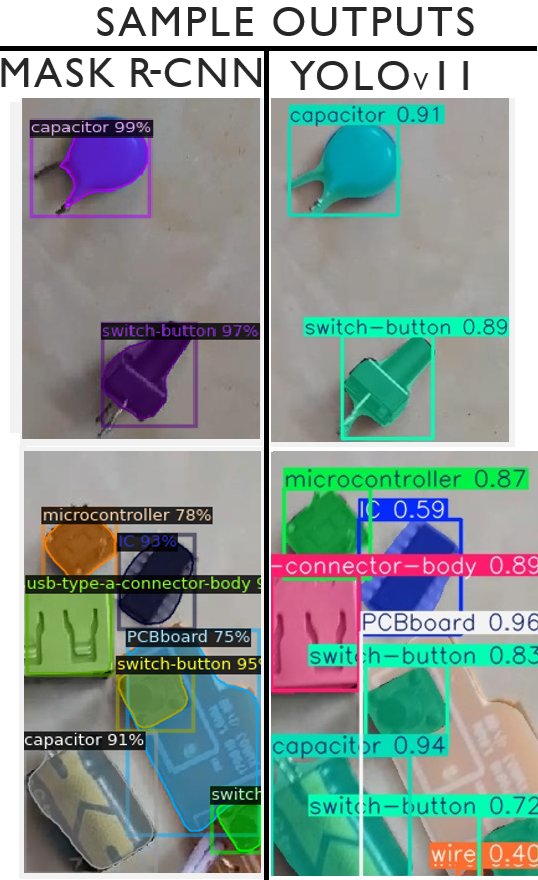}}
\caption{Comparison of sample outputs from Mask R-CNN model on the left and YOLOv11 model on the right.}
\label{fig}
\end{figure}

\begin{figure}[htbp]
\centerline{\includegraphics[width=0.99\linewidth]{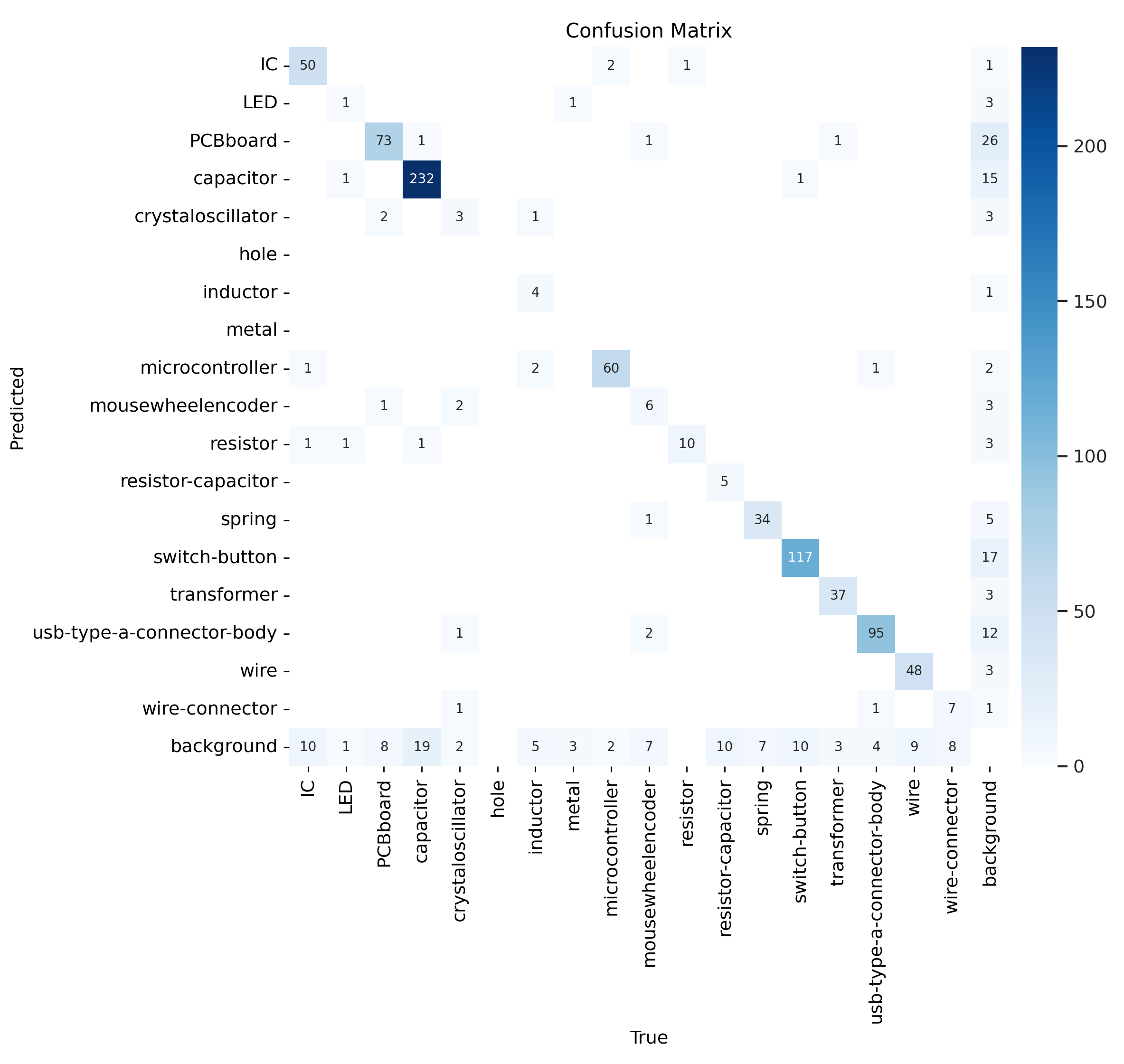}}
\caption{Confusion matrix of YOLOv11.}
\label{fig1}
\end{figure}

Fig.~\ref{fig} compares sample outputs between Mask R-CNN and YOLOv11 models. It can be seen that YOLOv11 is better at capturing finer details like thin wires, and in general, it has a slightly bigger mask than Mask R-CNN. The Mask R-CNN model took 3 times longer than YOLOv11 to train and couldn't identify objects from classes with fewer instances, like LED and mouse-wheel-encoder. 
Tables ~\ref{tab1} and ~\ref{tab2} show that YOLOv11 is much better than Mask R-CNN across all metrics. Fig.~\ref{fig1} shows the confusion matrix of the YOLOv11 model across all the classes of components.

\section{Conclusion and Future scope of work}
First, the dataset was created by taking pictures and a video of unsoldered components of a mouse and a charger. 100 pictures from the internet were included as well. The dataset size was increased from 643 to 6180 by applying augmentations and pre-processing techniques.

Second, two models: YOLOv11 and Mask R-CNN were applied and compared against various metrics.

YOLOv11 took around 1 hour to train while Mask R-CNN took 3 hours. Thus, it can be concluded that YOLOv11 is better than Mask R-CNN across all metrics.

This dataset can be extended to cover other classes of e-waste appliances in the future. Also, this computer vision model can be integrated with robotic software for real-world applications in industry.
There is a need to create an extensive dataset of e-waste in recycling facilities after it is crushed so that models to segregate it effectively can be created. 

\section*{Acknowledgment}

The author thanks Professor Rakesh Lingam, Theertha Biju, and Maniram Thota from the mechanical department of the Indian Institute of Technology Dharwad for their guidance, discussions, and help in this project.

\begin{table}[htbp]
\caption{Comparison of the two models for various metrics and Bounding Box Intersection over Union.}
\begin{center}
\begin{tabular}{|c|c|c|c|}
    \hline
        Model & R & mAP50 & mAP50:95 \\ \hline
        YOLOv11 & 65.5 & 71.6 &  58.5 \\ \hline
        Mask R-CNN& 38.4 &  41.1 & 26.9 \\ \hline
    \end{tabular}
\label{tab1}
\end{center}
\end{table}

\begin{table}[htbp]
\caption{Comparison of the two models for various metrics and Mask segmentation Intersection over Union.}
\begin{center}
\begin{tabular}{|c|c|c|c|}
    \hline
        Model & R & mAP50 & mAP50:95 \\\hline
        YOLOv11 & 65 & 70.7 &  51.6 \\\hline
        Mask R-CNN& 36.1 &  39.7 & 23.7 \\\hline
    \end{tabular}
\label{tab2}
\end{center}
\end{table}

\end{document}